\documentclass[a4paper,fleqn]{cas-dc}
\usepackage[numbers,sort&compress]{natbib}
\usepackage{tabularx}
\usepackage{xurl}
\makeatletter
\newcommand{\JVCIRStyleProbe}{%
\typeout{JVCIR-STYLE textwidth=\the\textwidth; textheight=\the\textheight; columnwidth=\the\columnwidth; columnsep=\the\columnsep}%
\typeout{JVCIR-STYLE fontsize=\f@size; baseline=\the\baselineskip; parindent=\the\parindent; parskip=\the\parskip; mathindent=\the\mathindent}%
\typeout{JVCIR-STYLE textfloatsep=\the\textfloatsep; dbltextfloatsep=\the\dbltextfloatsep; floatsep=\the\floatsep; intextsep=\the\intextsep}}
\makeatother

\begin{document}
\let\WriteBookmarks\relax
\shorttitle{Selective Surgical Action-Center Localization}
\shortauthors{Xue et al.}
\title[mode=title]{Selective Surgical Action-Center Localization: Retaining Localization Gains through Supervised Scoring}
\author[1]{Yanchong Xue}
\ead{thouofsgs@hrbeu.edu.cn}
\credit{Conceptualization, Methodology, Software, Validation, Formal analysis, Investigation, Data curation, Visualization, Writing -- original draft}
\author[1]{Wanting Zhang}
\ead{zhangwt@hrbeu.edu.cn}
\credit{Investigation, Validation}
\author[1]{Lingtao Yu}
\cormark[1]
\ead{yulingtaohrbeu@163.com}
\credit{Supervision, Resources}
\author[2]{Guochao Zhang}
\ead{m18689856730@163.com}
\credit{Investigation, Resources}
\affiliation[1]{organization={College of Mechanical and Electrical Engineering, Harbin Engineering University},
addressline={No. 145 Nantong Street, Nangang District},city={Harbin},postcode={150001},state={Heilongjiang Province},country={P. R. China}}
\affiliation[2]{organization={Department of Thoracic Surgery I, Hainan Cancer Hospital},
addressline={No. 6 Changbin West 4th Street, Xiuying District},city={Haikou},postcode={570312},state={Hainan Province},country={P. R. China}}
\cortext[1]{Corresponding author.}
\begin{abstract}
Surgical action-center localization requires both an accurate reference point and a decision about whether to report it. When frames with and without a reportable center are mixed, the benefit of improved localization depends on which correct candidates survive selection. We present a selective localization pipeline combining heatmap and coordinate supervision with a logistic-regression scorer. The scorer learns to identify predictions with a reportable center and a localization error within a specified tolerance. The internal evaluation uses 5,000 frames from 50 surgical sources, five source-grouped folds and two random seeds. Matched locator and scoring comparisons track correct candidates and their rejection. At an 80\% acceptance quota per source and a tolerance of 0.10 times the image diagonal, our method retains 46.08 correct outputs per 100 inputs. The baseline locator with the same scoring approach retains 44.20. The gain is 1.88 (descriptive 95\% source-paired bootstrap interval, 0.94 to 2.79). Under entropy-based selection, increased rejection of correct predictions offsets the higher supply of correct predictions on average. For the coordinate-supervised locator, supervised scoring reduces these rejections from 6.15 to 2.99 per 100 inputs. Evaluating retained correct outputs reveals whether a selection rule preserves or erodes an improvement in localization.
\end{abstract}

\maketitle
\hypersetup{pdfauthor={Yanchong Xue, Wanting Zhang, Lingtao Yu, Guochao Zhang},pdftitle={Selective Surgical Action-Center Localization: Retaining Localization Gains through Supervised Scoring}}
\JVCIRStyleProbe
% source block 4
\section{Introduction}\label{sec:1}

\begin{figure}[pos=tbp,width=\columnwidth]
\centering
\includegraphics[pagebox=cropbox]{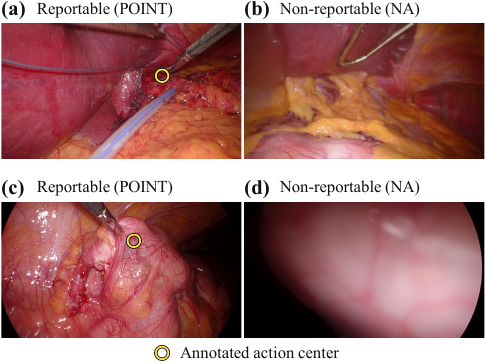}
\caption{Mixed reportability in surgical action-center localization. (a, c) POINT frames with an annotated action center, indicated by a yellow ring. (b, d) NA frames without a reportable center in the dataset annotation. Each row shows frames from the same surgical source at different times.}
\label{fig:1}
\end{figure}

% source block 5
Surgical video analysis provides information about operative activity for computer-assisted surgery \cite{ref01,ref02}. Camera assistance adds a spatial requirement: the region of interest must be located within the field of view \cite{ref03}. Surgical action-center localization estimates a task-related reference point whose displacement from the image center describes camera framing. The center of surgical action (COSA) dataset released by Ebrahimzadeh et al.\footnote{A. Ebrahimzadeh, N. Esmaeili, P. Horvath, M. Ghadimi and J. Hagenah, CDS-COSA-Dataset-2026, Zenodo, 2026. \url{https://doi.org/10.5281/zenodo.20627949}} provides point annotations for studying this task. Within a surgical source, frames with an annotated center (POINT) occur alongside frames annotated as non-reportable (NA) (Fig.~\ref{fig:1}). A localization system must therefore decide whether to report each predicted point.

% source block 6
An accepted point is correct when a reportable center exists and the predicted coordinate lies within a specified distance tolerance. Errors arise either from reporting a coordinate for a frame without a valid center or from placing it too far from an existing center. We therefore retain both annotation states and fix the acceptance budget, defined as the fraction of inputs retained. Counting correct outputs over all inputs evaluates reportability and localization accuracy together.

% source block 9
Heatmap learning and coordinate supervision provide complementary ways to train a point locator. Heatmaps retain the spatial organization of the image, while coordinate losses directly penalize displacement of the predicted point. Integral regression connects these representations through a differentiable spatial expectation \cite{ref04}, and subsequent work examines bias in this coordinate readout \cite{ref05}. Coordinate errors can thus guide learning through the heatmap. Related work on facial landmarks has also combined location estimation with uncertainty and visibility prediction, as in LUVLi \cite{ref06}.

% source block 10
Output selection requires a score that orders predictions by their suitability for acceptance. Selective prediction formalizes the trade-off between the fraction of predictions retained and their error \cite{ref07,ref08}. Medical image quality-control research likewise evaluates confidence scores through the predictions they retain or prioritize for correction \cite{ref09,ref10}. Comparisons must account for prediction quality as well as confidence ranking, with metrics aligned to the intended acceptance decision \cite{ref09,ref11}. For surgical action centers, we therefore ask whether a larger supply of correct coordinates translates into more correct reported points. At a common acceptance budget, the answer depends on both that supply and the correct candidates rejected by the scoring rule.

% source block 11
We present a selective surgical action-center localization pipeline that combines coordinate-supervised localization with learned success scoring. The locator uses heatmap and coordinate supervision with an expectation-based readout to generate a candidate point. A separately fitted logistic-regression scorer then uses image features and prediction descriptors to rank these candidates. The training target marks a prediction as successful when its frame is reportable and its coordinate meets the distance tolerance. The scorer is trained on predictions from the frozen locator. At the prescribed acceptance quota, the highest-scoring candidates within each source are retained, and the remaining points are withheld.

% source block 12
Our evaluation examines these two stages through controlled comparisons on 5,000 original frames from 50 surgical sources, with five source-grouped folds and two random seeds. We compare locators under the same scoring approach and compare scoring approaches with the locator fixed. This design relates output changes to correct candidate supply and selection losses. Entropy-based selection rejects enough additional correct predictions to offset the improved locator's larger supply of correct points on average. Supervised scoring retains a positive output gain at the same acceptance quota. Source-paired analysis and alternative scoring comparisons characterize the gain, while fixed development thresholds reveal how acceptance varies across sources.

% source block 13
Our contributions are twofold.

\begin{enumerate}

% source block 14
\item We develop a selective localization pipeline for surgical action centers in mixed reportable and non-reportable frames. The pipeline combines coordinate supervision with a scorer trained on the joint reportability and accuracy criterion, aligning selection with the output being evaluated.

% source block 15
\item We quantify how correct candidate supply and selection losses determine the output gain. Matched comparisons identify a mismatch between improved localization and entropy-based selection. Supervised scoring retains a positive output gain at the same acceptance budget.

\end{enumerate}

% source block 16
\section{Related work}\label{sec:2}

% source block 17
\subsection{Surgical action-center and point localization}\label{sec:2.1}

% source block 18; allow a column break at a hyphenated line.
{\brokenpenalty=100
EndoNet jointly learns surgical phases and instrument presence from laparoscopic videos \cite{ref12}. Rendezvous recognizes instrument--verb--target triplets, and Cholec\-Trip\-let2021 benchmarks this fine-grained description of surgical actions \cite{ref13,ref14}. Camera assistance requires a spatial reference within the image \cite{ref03}. The COSA dataset addresses this need through action-center coordinates and NA annotations. We use these annotations to evaluate selective point localization, retaining the original mixture of reportable and non-reportable frames.\par}

% source block 19
Point localization depends on the coordinate representation and its readout. DARK decodes heatmaps through a distribution-aware coordinate representation \cite{ref15}. Unbiased data processing addresses errors introduced by coordinate transformations and representation \cite{ref16}. Residual log-likelihood estimation learns an output distribution for coordinate regression \cite{ref17}, whereas SimCC classifies the horizontal and vertical coordinates separately \cite{ref18}. ViTPose uses a vision transformer with a lightweight pose decoder \cite{ref19}. These methods address coordinate estimation before any decision about reporting the point.

% source block 20
Sun et al. combine heatmap and coordinate regression through an integral operation \cite{ref04}. Gu et al. analyze and compensate for bias in integral regression \cite{ref05}. Our locator combines heatmap and coordinate supervision with an expectation over the full normalized sigmoid heatmap. We assess this training and readout combination by counting correct candidates before selection and correct outputs after selection under matched scoring approaches.

% source block 21
\subsection{Visibility and selective prediction}\label{sec:2.2}

% source block 22
Visibility modeling assigns a status to each landmark, while selective prediction governs whether a model reports its output. LUVLi jointly estimates facial landmark position, uncertainty and visibility through likelihood training \cite{ref06}. Gu et al. account for instance size and visibility, then learn network-specific confidence adjustments with a post-hoc Calibrated ConfidenceNet (CCNet) \cite{ref20}. In our surgical setting, reportability refers to the annotated presence of a valid action center. The success target combines that status with the predicted point meeting a distance tolerance.

% source block 23
Uncertainty estimation offers several sources of reliability information, including data uncertainty and uncertainty in the fitted model \cite{ref21,ref22}. Calibration evaluates how confidence agrees with predictive accuracy \cite{ref23}, while dataset shift can change the quality of uncertainty estimates \cite{ref24}. Output selection evaluates these scores through the errors remaining after low-scoring predictions are rejected. SelectiveNet learns prediction and selection jointly under a coverage constraint \cite{ref08}. Studies of failure detection and pretrained classifiers also show that the predictor, training regime and evaluation protocol affect confidence comparisons \cite{ref25,ref26}. For action-center localization, we fit the scorer to a frozen locator and compare accepted coordinates at common quotas.

% source block 24
\subsection{Quality scoring and output selection}\label{sec:2.3}

% source block 25
Medical image quality control connects prediction uncertainty to the quality of a complete output. Mehrtash et al. study calibration and predictive uncertainty in medical image segmentation \cite{ref27}. Geißler et al. show that structure size can confound correlations between uncertainty and segmentation quality \cite{ref28}. Zenk et al. compare failure-detection methods with risk--coverage analysis and identify confidence aggregation as a major factor in performance \cite{ref09}. SegQC predicts voxel-level errors and derives quality estimates that can prioritize corrections \cite{ref10}.

% source block 26
Uncertainty-aware quality prediction combines uncertainty maps with predicted segmentations and, in one formulation, the input image \cite{ref29}. Wundram et al. use split conformal prediction to estimate ranges of segmentation performance \cite{ref30}. Fournel et al. estimate image-specific segmentation difficulty from inter-observer agreement to inform quality thresholds \cite{ref31}. Our target is a reported action center that satisfies both reportability and coordinate accuracy.

% source block 27
Supervised confidence models estimate output quality directly from prediction features. Corbière et al. learn an estimate of true-class probability for failure prediction \cite{ref32}, and CCNet learns adjustments specific to a fitted pose estimator \cite{ref20}. Our scorer combines image and heatmap features with the candidate coordinate and a reportability cue. We compare locators and scores through rejected correct predictions and the two accepted-error categories, following task-specific evaluation of selective systems \cite{ref11}.

% source block 28
\section{Selective action-center localization}\label{sec:3}

\begin{figure*}[pos=tbp,width=\textwidth]
\centering
\includegraphics[pagebox=cropbox]{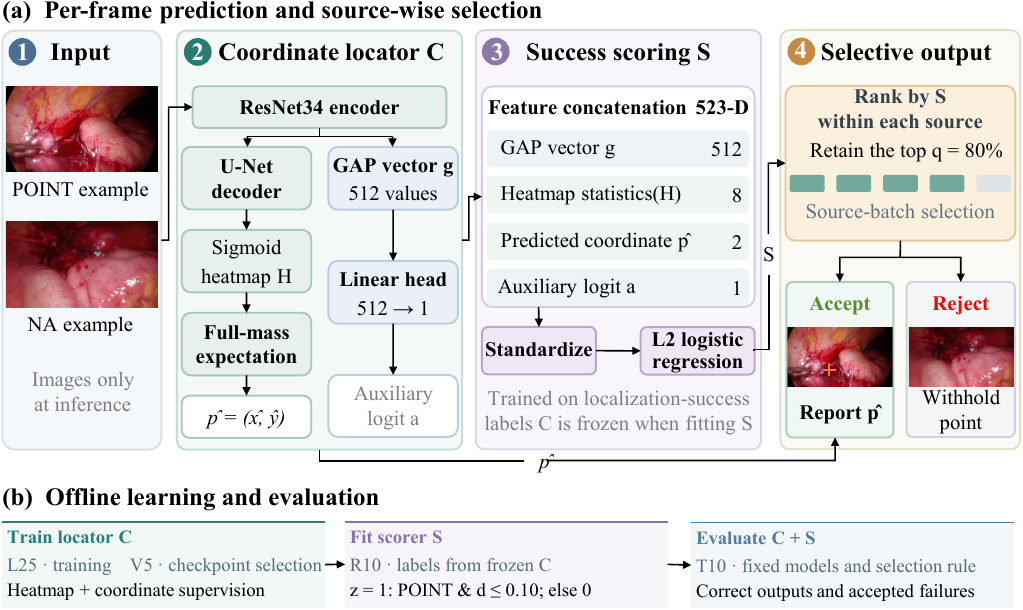}
\caption{Selective surgical action-center localization. (a) The locator C generates a sigmoid heatmap, an expectation-based coordinate, a global image representation and an auxiliary reportability logit. The 523-dimensional feature vector is standardized and passed to the logistic-regression scorer S. GAP denotes global average pooling, and L2 denotes the squared-weight regularization penalty. Candidates are ranked within each source, and the top 80\% are reported at the primary operating point. (b) Offline data roles separate locator training (L25), checkpoint selection (V5), scorer fitting (R10) and evaluation (T10). Numbers indicate sources per role. C is fixed when fitting S; labels identify POINT predictions within the 0.10-diagonal tolerance.}
\label{fig:2}
\end{figure*}

% source block 29
The C+S pipeline combines a coordinate-supervised locator C with a learned success scorer S (Fig.~\ref{fig:2}). For each color frame with red, green and blue (RGB) channels, C produces a candidate action center and features describing the image and prediction. S uses these features to score whether the candidate satisfies the localization success criterion. The selection rule then reports accepted coordinates and withholds the remaining points.

% source block 32
\subsection{Task and output criterion}\label{sec:3.1}

% source block 33
Each input ${x}_{i}$ is an RGB surgical frame with original width $W=1920$ and height ${H}_{0}=1080$ pixels. The annotation state ${r}_{i}\in \{0,1\}$ equals zero for a non-reportable frame (NA) and one for a frame with a reportable action center (POINT). A reference coordinate ${\mathbf{p}}_{i}\in {\mathbb{R}}^{2}$ is available for POINT frames. At inference, the pipeline receives the image and computes a candidate ${\hat{\mathbf{p}}}_{i}$ for every input in this mixed population.

% source block 34
We measure POINT localization error as the Euclidean distance from the reference coordinate, normalized by the original-image diagonal:

% source block 35; OMML index 6
\begin{equation}\label{eq:1}
d_i=\frac{\lVert\hat{\mathbf p}_i-\mathbf p_i\rVert_2}{\sqrt{W^2+H_0^2}},\quad r_i=1.
\end{equation}

% source block 36
Here the subscript 2 denotes the Euclidean norm. The success label combines reportability with a distance tolerance $\tau $:

% source block 37; OMML index 8
\begin{equation}\label{eq:2}
z_i(\tau)=\begin{cases}1,&r_i=1\ \mathrm{and}\ d_i\leq\tau,\\0,&\mathrm{otherwise}.\end{cases}
\end{equation}

% source block 38
The primary tolerance is $\tau =0.10$. NA frames receive ${z}_{i}=0$ directly from their annotation state. Let ${\delta }_{i}\in \{0,1\}$ denote rejection (zero) or acceptance (one). An accepted input returns its candidate coordinate ${\hat{\mathbf{p}}}_{i}$; a rejected input returns no point. A correct reported output therefore requires both ${\delta }_{i}=1$ and ${z}_{i}=1$.

% source block 39
\subsection{Coordinate-supervised localization}\label{sec:3.2}

% source block 40
The locator operates on a $512\times 288$ image canvas. A 34-layer residual network (ResNet34) encoder \cite{ref33} supplies multiscale features to a custom U-Net-style decoder \cite{ref34}. The five encoder feature maps have 64, 64, 128, 256 and 512 channels. Each feature map is projected to 64 channels. Four decoder stages successively upsample the coarser representation and concatenate the corresponding skip feature. Each stage applies two $3\times 3$ convolutions with eight-group normalization \cite{ref35} and rectified linear unit (ReLU) activation. A $1\times 1$ projection, bilinear resizing and sigmoid activation produce the single-channel heatmap ${H}_{i}$ at canvas resolution. Global average pooling of the final encoder map gives the 512-dimensional vector ${\mathbf{g}}_{i}$.

% source block 41
We recover coordinates through spatial expectation, following integral coordinate regression \cite{ref04,ref05}. Let $\Omega =\{0,\ldots ,w-1\}\times \{0,\ldots ,h-1\}$ denote the pixel grid, with width $w=512$ and height $h=288$. We normalize the sigmoid heatmap by its total mass and map the expected grid position to the original image:

% source block 42; OMML index 23
\begin{equation}\label{eq:3}
\begin{aligned}\pi_i(\mathbf u)&=\frac{H_i(\mathbf u)}{\sum_{\mathbf v\in\Omega}H_i(\mathbf v)},\\\hat{\mathbf p}_i&=\biggl(\frac Ww\sum_{\mathbf u\in\Omega}\pi_i(\mathbf u)u_x,\\&\qquad\frac{H_0}{h}\sum_{\mathbf u\in\Omega}\pi_i(\mathbf u)u_y\biggr)^T.\end{aligned}
\end{equation}

% source block 43
Here ${\pi }_{i}(\mathbf{u})$ is the normalized heatmap mass at pixel $\mathbf{u}=({u}_{x},{u}_{y})$, and $\mathbf{v}$ indexes the same grid. Superscript $T$ denotes transpose. The readout uses the full grid and passes coordinate gradients through the normalized heatmap. For each POINT frame, the target ${T}_{i}$ is a unit-peak Gaussian centered at the reference coordinate on the canvas, with standard deviation 32 pixels. Heatmap supervision constrains the spatial response, while coordinate supervision penalizes displacement of the reported candidate. For a training batch $\mathcal{B}$, the per-POINT localization loss and total objective are defined as follows, with vertical bars denoting set size:

% source block 44; OMML index 30
\begin{equation}\label{eq:4}
\begin{aligned}\ell_i^{\mathrm{loc}}&=\frac1{|\Omega|}\sum_{\mathbf u\in\Omega}(H_i(\mathbf u)-T_i(\mathbf u))^2\\&\quad+0.10d_i,\\\mathcal L_C&=\frac{\omega_f}{|\mathcal B|}\sum_{i\in\mathcal B:r_i=1}\ell_i^{\mathrm{loc}}\\&\quad+\frac{0.01}{|\mathcal B|}\sum_{i\in\mathcal B}\ell_{\mathrm{BCE}}(a_i,r_i).\end{aligned}
\end{equation}

% source block 45
Here ${\omega }_{f}={N}_{L,f}/{P}_{L,f}$ compensates for the POINT proportion in the locator-training split of fold $f$. ${N}_{L,f}$ counts all training inputs and ${P}_{L,f}$ counts its POINT frames. The denominator $|\mathcal{B}|$ is the actual batch size, including the four-frame final batch after full batches of eight. The auxiliary term uses binary cross-entropy (BCE) with logits to learn reportability from all training frames. Its logit, the raw reportability score, is ${a}_{i}={\mathbf{w}}_{a}^{T}\mathrm{stopgrad}({\mathbf{g}}_{i})+{b}_{a}$, with learned weights ${\mathbf{w}}_{a}$ and bias ${b}_{a}$. The stop-gradient operator prevents the auxiliary loss from updating the encoder. POINT frames provide localization supervision, while both annotation states train the auxiliary linear head. The resulting logit supplies one reportability cue to the success scorer.

% source block 46
\subsection{Learned success scoring}\label{sec:3.3}

% source block 47
After selecting the locator checkpoint, we freeze C and fit S to its predictions on the scoring-development sources. The offline roles in Fig.~\ref{fig:2}(b) assign disjoint source groups to locator training (L), checkpoint selection (V), scorer development (R) and evaluation (T). The fixed locator produces the R features and coordinates from which success labels are derived. The score uses scene information together with descriptors of the spatial prediction:

% source block 48; OMML index 39
\begin{equation}\label{eq:5}
\begin{aligned}\mathbf f_i&=\bigl[\mathbf g_i^T,\boldsymbol\phi(H_i)^T,\\&\qquad\hat x_i/W,\hat y_i/H_0,a_i\bigr]^T\in\mathbb R^{523},\\S_i&=\sigma\left(\boldsymbol\beta^T\widetilde{\mathbf f}_i+b\right).\end{aligned}
\end{equation}

% source block 49
The coordinates ${\hat{x}}_{i}$ and ${\hat{y}}_{i}$ in Eq.~\eqref{eq:5} are the horizontal and vertical components of the candidate ${\hat{\mathbf{p}}}_{i}$. The eight entries of $\boldsymbol{\phi }({H}_{i})$ are the normalized argmax coordinates, heatmap peak, heatmap mean, normalized entropy, two spatial variances and top-1\% mass. Argmax returns the heatmap peak location, with ties resolved by the first maximum in row-major order. The peak position is divided by $(w,h)$; the two additional coordinate entries in ${\mathbf{f}}_{i}$ describe the actual expectation-based output. Entropy is $-\sum_{\mathbf{u}} {\pi }_{i}(\mathbf{u})log[max\{{\pi }_{i}(\mathbf{u}),\varepsilon \}]/log|\Omega |$, with $\varepsilon $ set to the smallest positive normal 32-bit floating-point value. The variances are mass-weighted central second moments of ${u}_{x}/w$ and ${u}_{y}/h$. Top-1\% mass sums the largest 1,475 values of ${\pi }_{i}$. Combined with ${\mathbf{g}}_{i}$ and ${a}_{i}$, these descriptors provide S with image context and spatial prediction information.

% source block 50
Each feature is standardized using the mean and population standard deviation of the current fitting data. Columns with exactly zero standard deviation use a scale of one. The standardized vector ${\widetilde{\mathbf{f}}}_{i}$ enters the logistic model in Eq.~\eqref{eq:5}, where $\sigma $ is the sigmoid function. The vector $\boldsymbol{\beta }$ contains the learned feature weights, and $b$ is the intercept. We train on ${z}_{i}={z}_{i}(0.10)$ with an objective that gives each fitting source equal weight:

% source block 51; OMML index 58
\begin{equation}\label{eq:6}
\begin{aligned}\mathcal L_S&=\frac1{|\mathcal F|}\sum_{s\in\mathcal F}\frac1{n_s}\sum_{i\in\mathcal R_s}\ell_{\mathrm{CE}}(z_i,S_i)\\&\quad+\frac\lambda2\lVert\boldsymbol\beta\rVert_2^2.\end{aligned}
\end{equation}

% source block 52; local line-break tolerance prevents a 1.1 pt overrun.
{\emergencystretch=5pt
Here $\mathcal{F}$ is the set of fitting sources, and ${\mathcal{R}}_{s}$ contains the ${n}_{s}$ frames from source $s$. The cross-entropy is ${\ell }_{\mathrm{CE}}(z,S)=-z\mathrm{log}S-(1-z)\mathrm{log}(1-S)$; the intercept $b$ is unpenalized. We minimize this objective with the limited-memory, bound-con\-strained qua\-si-Newton optimizer L-BFGS-B \cite{ref36}. Leave-one-source-out development on the ten R sources selects the regularization strength $\lambda $ from $\{0.01,1,100\}$ by source-averaged cross-entropy. Each R source is held out in turn, with fitting and standardization performed on the remaining nine sources. For each locator, outer fold and training seed, we fit the final scorer on all ten R sources. Scorer training uses success labels at tolerance 0.10 throughout the evaluation. At inference, the fitted standardization parameters and logistic model are applied to the image-derived features.\par}

% source block 53
\subsection{Output selection}\label{sec:3.4}

% source block 54
The primary selection rule applies an acceptance quota within each source. For a source containing $n$ inputs and an acceptance fraction $q$, we rank candidates by descending ${S}_{i}$ and retain the first $\lfloor qn\rfloor $. The floor operator rounds the accepted count down to an integer. Ties are resolved by sorting sample identifiers in ascending character-code order. The primary quota $q=0.80$ therefore accepts 80 candidates from each 100-frame source. Coordinates and scores are computed per frame, while ranking uses all scores in the source batch. Acceptance returns the coordinate already produced by C, while rejection withholds that coordinate.

% source block 55
We also evaluate a fixed-threshold interface, with ${\delta }_{i}=\mathbf{1}[{S}_{i}\geq t]$. The bracketed indicator equals one when its condition holds and zero otherwise. The threshold $t$ is set from development scores and then held fixed for evaluation. Each acceptance decision then uses the current frame's score and the stored threshold. Its acceptance fraction can vary across sources.

% source block 56
\section{Experimental design}\label{sec:4}

% source block 57
We compared the co\-or\-di\-nate-supervised locator C with a controlled baseline B using the same scoring approach S and acceptance quota. B used heatmap mean-squared error (MSE) supervision and argmax readout, which returns the heatmap peak location. Entropy selection and alternative learned scores tested how the scoring rule and its training target changed the accepted set.

% source block 58
\subsection{Data and source roles}\label{sec:4.1}

% source block 59
The evaluation used 5,000 original annotated frames from the COSA dataset, comprising 100 frames from each of 50 sources. Each source groups ten clips from one surgery, with ten annotated keyframes per clip. Of these frames, 3,547 had a reportable action center (POINT) and 1,453 were labelled NA. The dataset uses NA when an action center is absent or ambiguous. All original annotated frames contributed to the evaluation, preserving the POINT/NA mixture. Images at $1920\times 1080$ pixels were mapped to the $512\times 288$ canvas described in Section~\ref{sec:3}, and coordinate errors were measured in the original image frame.

% source block 60
We used five source-grouped folds and two training seeds, 20260901 and 20260902. Within each fold, disjoint sets of 25, 5, 10 and 10 sources served as L, V, R and T, respectively (Fig.~\ref{fig:2}). L trained the locator, V selected its checkpoint, R supported score fitting and threshold development, and T supplied evaluation outputs. Each source entered T in one fold. Grouping by source kept related frames in the same data role within a fold, following methodological recommendations \cite{ref37,ref38}. Prior analyses of the same source collection informed the choice of C+S and the scoring comparisons.

% source block 61
\subsection{Training and scoring protocol}\label{sec:4.2}

% source block 62
B and C used the same ResNet34-based architecture, ImageNet \cite{ref39} encoder initialization and 60-epoch training budget. For each seed, their sample order and horizontal flips were matched. Images used ImageNet channel normalization, and horizontal flips were applied with probability 0.5. Both encoders were fine-tuned throughout training with frozen batch-normalization \cite{ref40} running statistics. We used 32-bit floating-point (FP32) computation, disabled TensorFloat-32 (TF32) and trained with batches of eight, retaining the final partial batch.

% source block 63
The AdamW optimizer with decoupled weight decay \cite{ref41} used learning rates of ${10}^{-5}$ for the encoder and ${10}^{-4}$ for the decoder and auxiliary head. Weight decay was ${10}^{-4}$ with moment-decay coefficients $({\beta }_{1},{\beta }_{2})=(0.9,0.999)$. At epoch $e\in \{1,\ldots ,60\}$, both learning rates were multiplied by $0.01+0.99[1+\mathrm{cos}(\pi (e-1)/59)]/2$. Checkpoint selection maximized the source-macro (equal-source average) proportion of POINT frames within tolerance 0.10 on V. Ties favored lower mean normalized distance, then the earlier epoch. The auxiliary head and its supervision were present in both B and C.

% source block 64
We froze each locator and fitted its scoring models separately for every fold and seed. The R10 source-leave-one-out procedure in Section~\ref{sec:3.3} selected regularization from $\{0.01,1,100\}$ by source-macro cross-entropy, with stronger regularization resolving ties. Each fitting split supplied its own standardization statistics. A final fit on all R10 sources produced the scorer applied to T. The reportability and alternative score models followed the same development procedure for their respective targets. All analyses used the saved fitted scores, with S trained at tolerance 0.10 throughout.

% source block 65
\subsection{Comparisons and metrics}\label{sec:4.3}

% source block 66
Comparing C+S with B+S tested the complete co\-or\-di\-nate-supervised training and readout combination under a shared scoring approach. Comparing C+S with C+E held the coordinates fixed and changed output selection, where E ranked lower normalized heatmap entropy first. B+E completed the four configurations. We also evaluated J, which allowed the auxiliary loss to update the shared encoder in the C architecture, and a SegFormer-MiT-B1 locator \cite{ref42} adapted to the same data roles, MSE supervision, argmax readout and training budget. Both used separately fitted S scores.

% source block 67
Alongside E and S, the scoring study evaluated heatmap peak K and five further scores. The reportability score ${p}_{R}$ predicted POINT from the same 523 features as S. GateK placed candidates with ${p}_{R}\geq 0.5$ first and ranked by peak within each group. F used the successful-localization class score from a three-state model for NA, POINT excess error and POINT success. The factorized score $P{Q}_{R}={p}_{R}{s}_{\mathrm{cond}}$ combined reportability with ${s}_{\mathrm{cond}}$, the predicted success probability conditional on a POINT frame. $P{Q}_{\mathrm{aux}}={p}_{\mathrm{aux}}{s}_{\mathrm{cond}}$ replaced ${p}_{R}$ with ${p}_{\mathrm{aux}}$, the reportability probability from a model fitted to the scalar auxiliary logit. Each target had its own selected regularization.

% source block 68
We measured correct points supplied and retained over the full input population, following problem-specific image-analysis validation principles \cite{ref43,ref44}. For a source with $n$ inputs, the success indicator ${z}_{i}(\tau )$ and acceptance indicator ${\delta }_{i}$ defined in Section~\ref{sec:3} give correct localization supply G, correct accepted outputs Y and rejected correct predictions M as

% source block 69; OMML index 93
\begin{equation}\label{eq:7}
\begin{aligned}G&=\frac{100}{n}\sum_{i=1}^n z_i,\\Y&=\frac{100}{n}\sum_{i=1}^n\delta_i z_i,\\M&=\frac{100}{n}\sum_{i=1}^n(1-\delta_i)z_i.\end{aligned}
\end{equation}

% source block 70
All three quantities are counts per 100 original inputs and satisfy $G=Y+M$. The change in Y between configurations is therefore the change in G minus the change in M. Coverage is the fraction of all inputs accepted; accepted risk is the failure fraction within the accepted set:

% source block 71; OMML index 95
\begin{equation}\label{eq:8}
\begin{aligned}\mathrm{coverage}&=\frac1n\sum_{i=1}^n\delta_i,\\\mathrm{risk}&=\frac{\sum_{i=1}^n\delta_i(1-z_i)}{\sum_{i=1}^n\delta_i}.\end{aligned}
\end{equation}

% source block 72
Risk is evaluated when at least one prediction is accepted. Accepted failures comprise non-reportable frames (NA) and POINT predictions beyond tolerance (PE). Severe errors (SE) are the subset of accepted POINT predictions with $d>0.20$. In count expressions, NA and PE denote the accepted counts in their respective categories. Using the same per-100-input units, $Y+\mathrm{NA}+\mathrm{PE}=100\mathrm{coverage}$. Thus, at a fixed quota, Y and risk express the same count relationship. Our primary setting was $\tau =0.10$ and $q=0.80$. We also evaluated quotas $q\in \{0,0.20,0.40,0.60,0.80,0.90,1\}$ and tolerances $\tau \in \{0.05,0.10,0.20\}$.

% source block 73
\subsection{Source-level analysis and threshold transfer}\label{sec:4.4}

% source block 74
For each source, we averaged its two seed-specific metrics and compared configurations using paired differences across the 50 equally weighted sources. Descriptive 95\% intervals used the 2.5th and 97.5th percentiles of 20,000 bootstrap resamples \cite{ref45} of the 50 paired source differences. Resampling used the PCG64 random-number generator with seed 20260903. These intervals describe variation among evaluated sources with the saved models and configuration choices held fixed. Folds shared training sources, while their T sources were disjoint. We also summarized the outputs separately by fold and by seed.

% source block 75
For fixed-threshold evaluation, each fold, seed and configuration received a numeric threshold from its 1,000 R development scores. The threshold was the 800th score in descending order, and all scores at or above it were accepted, including boundary ties. E used negative entropy so that larger scores indicated higher priority. S used out-of-fold development scores at the selected regularization. Each R source was scored by a model fitted to the other R sources. That threshold was then applied to T scores from the model refitted on all R10 sources. We reported actual coverage and source-macro risk, with each source and seed combination using its own acceptance count as the risk denominator.

% source block 76
\section{Results}\label{sec:5}

% source block 77
\subsection{Selective localization at a fixed acceptance budget}\label{sec:5.1}

% source block 78
At $\tau =0.10$ and $q=0.80$, C+S yielded 46.08 correct accepted outputs per 100 original inputs, compared with 44.20 for B+S (Table~\ref{tab:1}). The paired difference was +1.88, with a descriptive 95\% interval of [0.94, 2.79]. Both configurations accepted 80 predictions per source, directly comparing correctness at the same output budget. The corresponding accepted risk was 42.40\% for C+S and 44.75\% for B+S.

\begin{figure}[pos=tbp,width=\columnwidth]
\centering
\includegraphics[pagebox=cropbox]{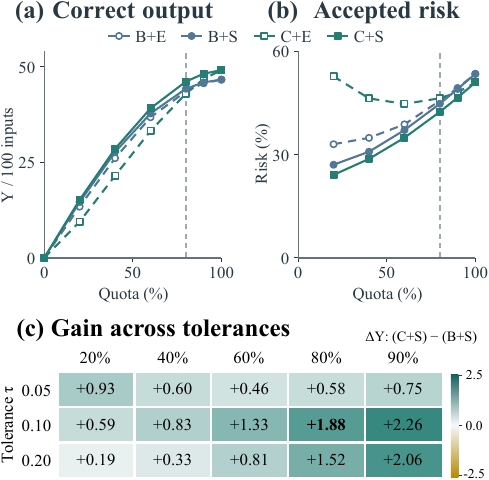}
\caption{Selective localization across acceptance budgets. (a) Correct accepted outputs Y per 100 original inputs and (b) accepted risk at $\tau =0.10$. The vertical dashed line marks the primary 80\% quota. Risk is undefined at zero acceptance. (c) Mean output differences between C+S and B+S across tolerances and quotas. Values average two seeds within each of 50 sources, followed by an equal-source average. C+S is our method; B and C denote the baseline and coordinate-supervised locators, and E and S denote entropy selection and learned success scoring.}
\label{fig:3}
\end{figure}

\begin{table}[pos=tbp,width=\columnwidth]
\caption{Selective localization at $\tau =0.10$ and $q=0.80$. G is correct localization supply, Y is correct accepted output and M is rejected correct predictions, all per 100 original inputs. Risk is the failure percentage among accepted outputs. Means weight the 50 sources equally after averaging their two seeds. B and C denote the baseline and coordinate-supervised locators; E and S denote entropy selection and learned success scoring. The lower panel reports source-paired differences in Y (\ensuremath{\Delta}Y) with descriptive 95\% bootstrap intervals. Bold identifies the proposed C+S configuration.}
\label{tab:1}
\begin{tabular*}{\tblwidth}{@{\extracolsep{\fill}}@{}lrrrr@{}}
\toprule
\textbf{Configuration} & \textbf{G \ensuremath{\uparrow}} & \textbf{Y \ensuremath{\uparrow}} & \textbf{M \ensuremath{\downarrow}} & \textbf{Risk (\%) \ensuremath{\downarrow}} \\
\midrule
B+E & 46.58 & 43.50 & 3.08 & 45.63 \\
B+S & 46.58 & 44.20 & 2.38 & 44.75 \\
C+E & 49.07 & 42.92 & 6.15 & 46.35 \\
\textbf{Ours (C+S)} & \textbf{49.07} & \textbf{46.08} & \textbf{2.99} & \textbf{42.40} \\
\midrule
\multicolumn{2}{l}{\textbf{Comparison}} & \textbf{\ensuremath{\Delta}Y \ensuremath{\uparrow}} & \multicolumn{2}{c}{\textbf{95\% interval}} \\
\midrule
\multicolumn{2}{l}{C+S \ensuremath{-} B+S} & +1.88 & \multicolumn{2}{c}{[0.94, 2.79]} \\
\multicolumn{2}{l}{C+S \ensuremath{-} C+E} & +3.16 & \multicolumn{2}{c}{[2.26, 4.11]} \\
\multicolumn{2}{l}{C+S \ensuremath{-} B+E} & +2.58 & \multicolumn{2}{c}{[1.55, 3.60]} \\
\bottomrule
\end{tabular*}
\end{table}

% source block 82
C+S retained a positive mean advantage over B+S across the evaluated nonzero acceptance budgets (Fig.~\ref{fig:3}). With S, replacing B by C increased Y by 0.59, 0.83, 1.33, 1.88 and 2.26 at quotas of 20\%, 40\%, 60\%, 80\% and 90\%, respectively. The tolerance-by-quota matrix also showed positive mean differences at the stricter 0.05 and broader 0.20 tolerances. These evaluations used the same saved scores throughout. At full acceptance, all candidates were retained and the output count equalled the correct localization supply, 46.58 for B and 49.07 for C.

% source block 85
\subsection{Localization gains and selection losses}\label{sec:5.2}

% source block 86
Under entropy selection, additional rejections of correct predictions exceeded the increase in correct localization supply on average. Moving from B to C increased G by 2.49 per 100 inputs, but increased M from 3.08 to 6.15 (Fig. 4a). The resulting change in Y was $2.49-3.07=-0.58$, with a descriptive interval of [\ensuremath{-}1.65, 0.51].

% source block 87
Supervised scoring preserved a positive net gain from the larger coordinate supply. Using the same scoring approach S, moving from B to C increased M by 0.61, leaving a net gain of $2.49-0.61=1.88$ in Y. Holding C fixed instead, replacing E with S reduced M from 6.15 to 2.99 and increased Y by 3.16 (descriptive 95\% interval, [2.26, 4.11]). The locator comparison measures the net output gain after both supply and rejection change. The scoring comparison measures the gain from selecting among a fixed set of coordinates.

% source block 88
The two comparisons changed the composition of accepted errors differently (Fig. 4b). Relative to B+S, C+S accepted 0.42 more NA frames and 2.30 fewer POINT predictions beyond tolerance per 100 inputs. Relative to C+E, it accepted 4.63 fewer NA frames and 1.47 more POINT predictions beyond tolerance.

\begin{figure}[pos=tbp,width=\columnwidth]
\centering
\includegraphics[pagebox=cropbox]{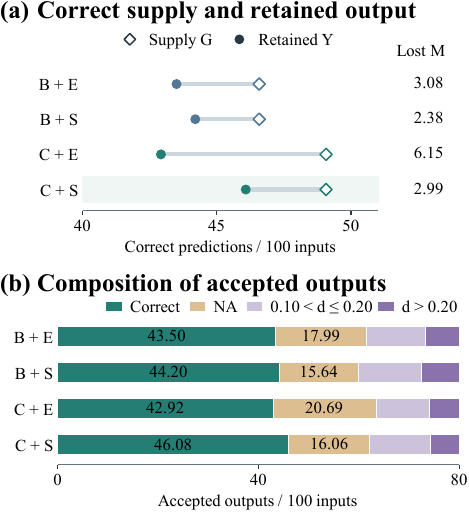}
\caption{Correct coordinate supply and selection outcomes at $\tau =0.10$ and $q=0.80$. (a) Open diamonds mark supply G, filled circles mark retained correct outputs Y, and the right-hand values give rejected correct predictions M. (b) Accepted outputs comprise correct predictions, NA frames, POINT errors with $0.10<d\leq 0.20$, and severe POINT errors with $d>0.20$. The four segments sum to 80 accepted outputs per 100 inputs. All values are source-macro means after averaging the two seeds within source.}
\label{fig:4}
\end{figure}

% source block 91
\subsection{Variation across sources}\label{sec:5.3}

% source block 92
C+S increased Y relative to B+S in 33 of 50 sources, matched it in four and decreased it in 13 (Fig. 5a). The five fold means were all positive, ranging from +0.45 to +3.05 per 100 inputs. The seed-specific mean differences were +2.40 and +1.36.

% source block 93
At fixed coordinates, S increased Y over E in 42 sources, matched it in two and decreased it in six (Fig. 5b). Fold means ranged from +1.25 to +5.50, and the two seed means were +2.90 and +3.42. More sources gained correct outputs from changing E to S on C than from changing B to C under S.

\begin{figure}[pos=tbp,width=\columnwidth]
\centering
\includegraphics[pagebox=cropbox]{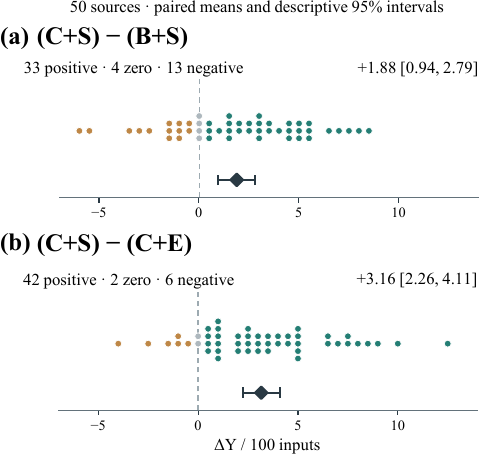}
\caption{Source-paired changes in correct accepted outputs at $\tau =0.10$ and $q=0.80$. Each dot represents one source after averaging its two seeds. (a) C+S minus B+S. (b) C+S minus C+E. Diamonds and horizontal bars show the mean and descriptive 95\% percentile-bootstrap interval from 20,000 source resamples. Positive, zero and negative source counts are stated above each distribution; vertical stacking separates coincident points.}
\label{fig:5}
\end{figure}

% source block 96
\subsection{Scoring strategies and locator configurations}\label{sec:5.4}

% source block 97; permit a natural column break without changing typography.
{\brokenpenalty=100
Learned scoring retained more correct outputs than entropy selection in configuration C (Fig. 6a). At the primary operating point, E yielded 42.92 correct outputs per 100 inputs and peak scoring K yielded 44.59. Adding reportability grouping in GateK increased this to 45.49. The learned scores clustered between 46.05 and 46.50: S and the three-state score F both yielded 46.08, $P{Q}_{R}$ yielded 46.15, and $P{Q}_{\mathrm{aux}}$ gave the highest mean of 46.50.\par}

% source block 98
Success and reportability supervision produced close mean output counts with the same feature schema. At 80\% acceptance, S exceeded ${p}_{R}$ by 0.03 [\ensuremath{-}0.36, 0.42] with C and by 0.21 [\ensuremath{-}0.16, 0.61] with B (Fig. 6c). Across quotas, their mean differences varied with the acceptance budget (Fig. 6b). For C at the primary point, S accepted 0.64 more NA frames and 0.67 fewer POINT excess errors than ${p}_{R}$. The close output counts therefore came from different accepted error compositions.

\begin{figure}[pos=tbp,width=\columnwidth]
\centering
\includegraphics[pagebox=cropbox]{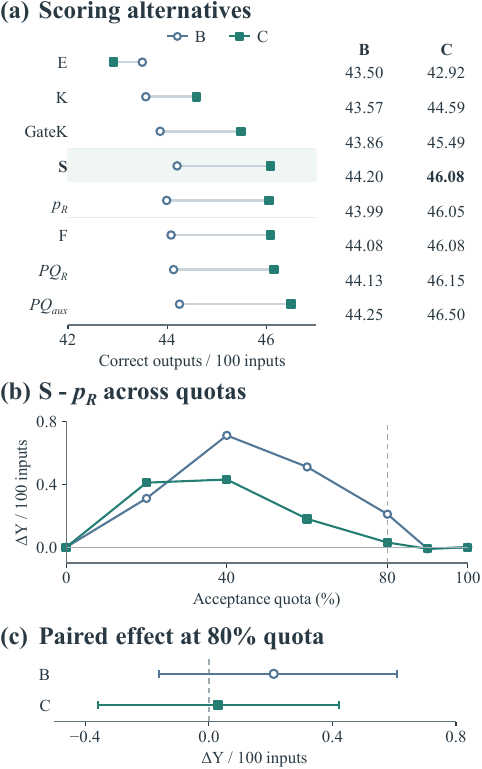}
\caption{Scoring comparisons at $\tau =0.10$. (a) Correct accepted outputs per 100 inputs for eight scores on B and C at $q=0.80$. Scores are defined in Section~\ref{sec:4.3}. (b) Mean difference in Y between success supervision S and reportability supervision ${p}_{R}$ across quotas. (c) Their source-paired mean differences and descriptive 95\% intervals at 80\% acceptance. S and ${p}_{R}$ use the same 523-feature schema with target-specific fitting. All summaries average seeds within source before aggregating 50 sources.}
\label{fig:6}
\end{figure}

% source block 101
C and J produced close mean output counts under S (Table~\ref{tab:2}). They yielded 46.08 and 45.96 correct outputs per 100 inputs, respectively, compared with 44.20 for B and 42.04 for the SegFormer adaptation. Their correct supplies were 49.07 and 48.40, while their rejected correct predictions were 2.99 and 2.44.

\begin{table}[pos=tbp,width=\columnwidth]
\caption{Locator configurations with separately fitted S at $\tau =0.10$ and $q=0.80$. G and Y are correct supply and correct accepted outputs. SE counts accepted POINT predictions with $d>0.20$. B is the baseline, C is the coordinate-supervised locator, and J also lets the auxiliary loss update the shared encoder. All values are per 100 original inputs and use the same source-macro aggregation.}
\label{tab:2}
\begin{tabular*}{\tblwidth}{@{\extracolsep{\fill}}@{}lrrr@{}}
\toprule
\textbf{Locator + S} & \textbf{G \ensuremath{\uparrow}} & \textbf{Y \ensuremath{\uparrow}} & \textbf{SE \ensuremath{\downarrow}} \\
\midrule
B+S & 46.58 & 44.20 & 7.60 \\
\textbf{Ours (C+S)} & \textbf{49.07} & \textbf{46.08} & \textbf{5.66} \\
J+S & 48.40 & 45.96 & 5.60 \\
SegFormer+S & 44.18 & 42.04 & 7.99 \\
\bottomrule
\end{tabular*}
\end{table}

% source block 105
\subsection{Frozen thresholds and computational cost}\label{sec:5.5}

% source block 106
Frozen development thresholds retained the output advantage of C+S at the resulting acceptance levels (Table~\ref{tab:3}). Our method achieved 78.97\% coverage and 46.85 correct outputs per 100 inputs, with source-macro accepted risk of 42.45\%. B+S achieved 79.23\% coverage and 44.70 correct outputs, whereas C+E achieved 78.37\% coverage and 42.83 correct outputs.

\begin{table}[pos=tbp,width=\columnwidth]
\caption{Frozen-threshold evaluation at $\tau =0.10$. Thresholds were developed at 80\% acceptance on R and applied numerically to T. Coverage is the actual accepted percentage, Y is per 100 original inputs, and risk is the source-macro failure percentage among accepted outputs. All 100 source-seed units had at least one accepted prediction.}
\label{tab:3}
\begin{tabular*}{\tblwidth}{@{\extracolsep{\fill}}@{}lrrr@{}}
\toprule
\textbf{Configuration} & \textbf{Coverage (\%)} & \textbf{Y \ensuremath{\uparrow}} & \textbf{Risk (\%) \ensuremath{\downarrow}} \\
\midrule
B+E & 79.48 & 44.18 & 45.64 \\
B+S & 79.23 & 44.70 & 45.09 \\
C+E & 78.37 & 42.83 & 46.30 \\
\textbf{Ours (C+S)} & \textbf{78.97} & \textbf{46.85} & \textbf{42.45} \\
\bottomrule
\end{tabular*}
\end{table}

\begin{figure}[pos=tbp,width=\columnwidth]
\centering
\includegraphics[pagebox=cropbox]{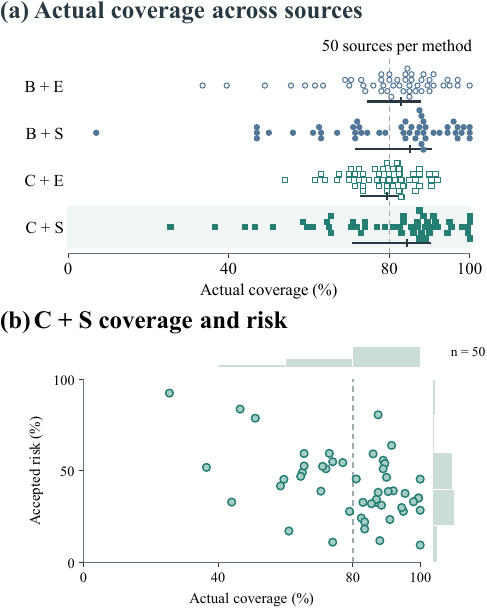}
\caption{Coverage after transferring frozen development thresholds to T, at $\tau =0.10$. (a) Source coverage for each configuration, with median ticks and interquartile-range bars. (b) Source coverage and accepted risk for C+S, with marginal source counts. Each point averages the two seeds within one source. The dashed line marks 80\%, the development acceptance target. Risk is calculated within each source and seed before averaging.}
\label{fig:7}
\end{figure}

% source block 110
Coverage varied substantially across sources (Fig.~\ref{fig:7}). After averaging seeds within each source, C+S coverage ranged from 25.5\% to 100\%, with a median of 84.5\%. The source-wise scatter in Fig. 7b also showed that similar acceptance fractions accompanied different accepted risks. Risk in Table~\ref{tab:3} averages the source-specific ratios defined in Eq.~\eqref{eq:8}.

% source block 111
On an NVIDIA GeForce RTX 5060 Ti, C+S processing took 22.60 ms per input at batch size one. At batch size eight, it took 15.90 ms per input. Both values averaged three timing blocks using fold 0 and seed 20260901. Timing started with preprocessing of decoded frames and covered device transfer, locator inference, coordinate readout, feature extraction and scoring. The locator and auxiliary head contained 21,794,498 parameters.

% source block 114
\subsection{Qualitative examples}\label{sec:5.6}

\begin{figure*}[pos=tbp,width=\textwidth]
\centering
\includegraphics[pagebox=cropbox]{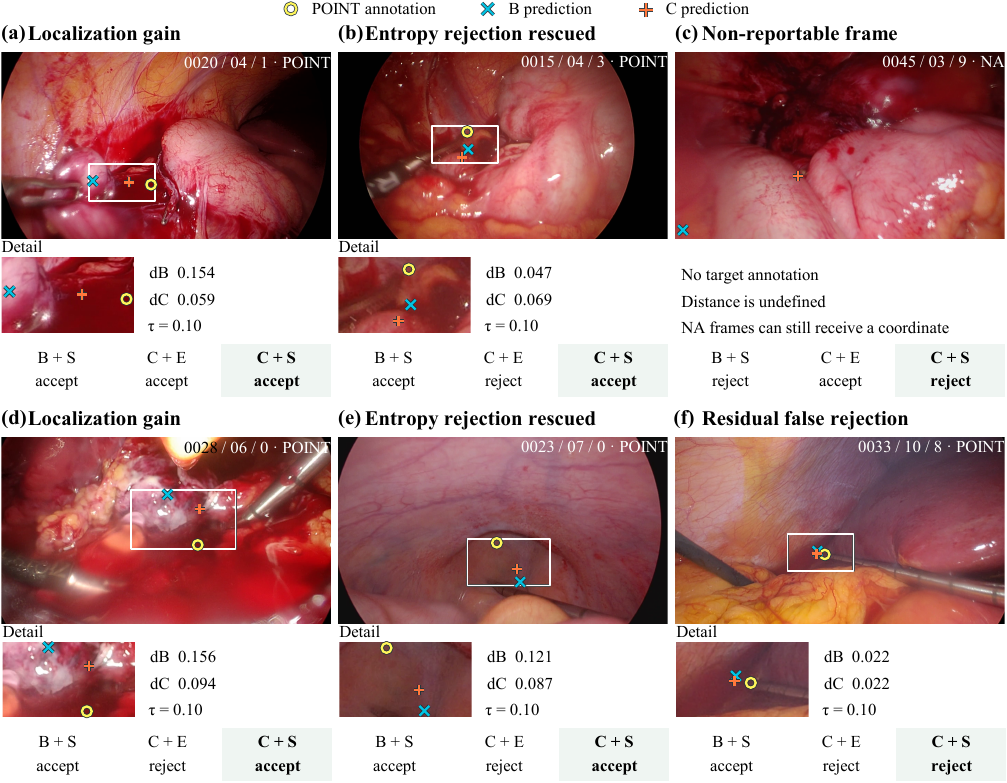}
\caption{Examples of localization and selection at $\tau =0.10$, $q=0.80$ and seed 20260901. Yellow circles denote POINT annotations, cyan crosses B predictions and orange plus signs C predictions. Panels a and d show coordinates entering tolerance; b and e show correct C coordinates retained by S after rejection by E; c shows NA rejection; f shows a residual correct rejection. Full frames retain the original view. White boxes locate the detail crops, whose extents vary to contain the annotation and both predictions; distances beside each crop use the original-image diagonal. The three decisions beneath each frame correspond to B+S, C+E and C+S.}
\label{fig:8}
\end{figure*}

% source block 115
The six examples in Fig.~\ref{fig:8} connect coordinate changes and acceptance decisions to individual frames. From B to C, normalized errors in panels a and d decreased from 0.154 to 0.059 and from 0.156 to 0.094, respectively. Both C predictions entered the 0.10 tolerance and were accepted by S. Panel d also illustrates the effect of selection: E rejected the correct C prediction that S retained.

% source block 116
Panels b and e show correct C predictions recovered from entropy rejection. In b, both locators were within tolerance, with errors of 0.047 for B and 0.069 for C. S retained the C coordinate rejected by E. In e, the error decreased from 0.121 to 0.087. The C coordinate entered tolerance and was again retained by S after rejection by E.

% source block 117
Panel c is an NA frame: C+E accepted a coordinate, whereas C+S withheld it. Panel f shows a correct C prediction with error 0.022 that both E and S rejected. This rejected correct candidate contributes to M in Table~\ref{tab:1}.

% source block 120
\section{Discussion}\label{sec:6}

% source block 121
C supplied more correct coordinates, but its output gain depended on the scoring rule (Fig.~\ref{fig:4}). Entropy selection rejected enough correct candidates to offset the increased supply on average, whereas supervised scoring preserved a positive net gain. Separating candidate supply from selection losses shows where these two outcomes diverged. A change in locator should therefore be assessed with its scoring rule, using retained correct outputs as the endpoint.

% source block 122
For action centers, selective prediction must account for both reportability and coordinate accuracy. SelectiveNet optimizes prediction quality on the retained population under a coverage constraint \cite{ref08}. Zenk et al. evaluate segmentation quality together with confidence ranking \cite{ref09}, and SegQC assesses the use of quality estimates to prioritize corrections \cite{ref10}. The mixed POINT/NA population exposes two kinds of accepted error: false reports on NA frames and misplaced POINT predictions. The error composition in Fig. 4b shows which type changes in each comparison, explaining how the gain in correct outputs arises.

% source block 123
S learns the same joint reportability and localization criterion used to count correct outputs. It combines image and prediction features to rank candidates for acceptance. With the same feature schema, S and reportability scoring ${p}_{R}$ produced close mean output counts (Fig.~\ref{fig:6}). Reportability cues may also carry information about localization difficulty, allowing either target to learn a useful ranking. Factorized scores combine reportability and coordinate success explicitly, with $P{Q}_{\mathrm{aux}}$ achieving the highest mean in the scoring comparison. On average, all of these learned scores retained more correct C predictions than entropy selection.

% source block 124
A quota fixes the number retained within a source batch and depends on score order. An order-preserving change in score scale leaves that selection intact but can change which frames cross a numeric threshold. Calibration and ranking therefore address different aspects of confidence use \cite{ref23,ref26}. Frozen thresholds produced mean coverage close to the development target, alongside a broad range of source-specific coverage (Fig.~\ref{fig:7}). Threshold transfer involves new sources and a change from source-leave-one-out fitting to a scorer fitted on all R sources. Both changes may affect the score scale, which matters alongside ranking quality when using a stored threshold. Source-specific coverage and accepted risk show how the stored threshold behaves after transfer.

% source block 125
For camera-reference generation, the distance tolerance specifies acceptable displacement and the acceptance budget sets the fraction of frames that return a reference. Choosing a locator and selection rule then requires considering how many references are correct and whether errors arise from false reports or misplaced coordinates.

% source block 126
\section{Conclusion}\label{sec:7}

% source block 127; permit a natural page break without changing typography.
{\brokenpenalty=100
We presented a selective localization pipeline for surgical action centers in frames with mixed reportability, combining coordinate-supervised localization with learned success scoring. Across the 50 surgical sources, our method retained 1.88 more correct outputs per 100 original inputs than the baseline locator with the same scoring approach. This gain was measured at an 80\% acceptance quota per source and a tolerance of 0.10 times the image diagonal. Matched comparisons separated the increase in correct candidate supply from losses during selection. Entropy-based rejection offset the larger supply on average, whereas supervised scoring preserved a positive net gain. Counting retained correct outputs measures how much of a localization gain survives selection.\par}

\printcredits
\section*{Funding}
The authors received no specific funding for this work.
\section*{Declaration of competing interest}
The authors declare no conflict of interest.
\section*{Data availability}
The original surgical clips and action-center annotations are available in the CDS-COSA-Dataset-2026 repository (\url{https://doi.org/10.5281/zenodo.20627949}), under the Creative Commons Attribution-NonCommercial 4.0 International licence. The aggregate results supporting the reported comparisons are presented in the figures and tables of this article.
\section*{Code availability}
The code supporting this study is being prepared for public release at \url{https://github.com/thouofsgs/selective-cosa-localization}.
\section*{Declaration of generative AI and AI-assisted technologies in the manuscript preparation process}
During the preparation of this work, the authors used ChatGPT (OpenAI) to assist with language editing and improve the readability of the manuscript. The authors reviewed and edited the resulting content as needed and take full responsibility for the content of the article.
\section*{Acknowledgements}
The authors have no acknowledgments to declare.

\end{document}